\documentclass{egpubl}
\usepackage{eurova2025}

 \WsPaper         

\usepackage[T1]{fontenc}
\usepackage{dfadobe}  
\usepackage{booktabs}                 
\usepackage{tabularx}
\usepackage{subcaption}
\usepackage{makecell}
\usepackage{multirow}
\usepackage{bm}
\usepackage{listings}
\usepackage[svgnames]{xcolor}
\usepackage[english]{babel}
\usepackage{cite} 
\BibtexOrBiblatex

\electronicVersion
\PrintedOrElectronic
\ifpdf \usepackage[pdftex]{graphicx} \pdfcompresslevel=9
\else \usepackage[dvips]{graphicx} \fi

\usepackage{egweblnk} 

\definecolor{colorIncluded}{RGB}{188, 221, 149}
\definecolor{colorDiscarded}{RGB}{230, 160, 155}

\addto\extrasenglish{
}

\title{Cutting Through the Clutter: The Potential of LLMs for Efficient Filtration in Systematic Literature Reviews}

\author[Joos et al.]
{\parbox{\textwidth}{\centering
    \vspace*{-5mm}
    Lucas Joos$^{1}$\orcid{0000-0001-7049-5203},
    Daniel A. Keim$^{1}$\orcid{0000-0001-7966-9740}, and
    Maximilian T. Fischer$^{1}$\orcid{0000-0001-8076-1376}
}
        \\
{\parbox{\textwidth}{\centering \vspace*{-5mm}
 $^1$University of Konstanz, Germany
       }
}
}

\begin{document}

\teaser{
    \vspace*{-7mm}
    \includegraphics[width=1\linewidth, trim={0 5.6cm 1.75cm 0},clip, alt={schematic}]{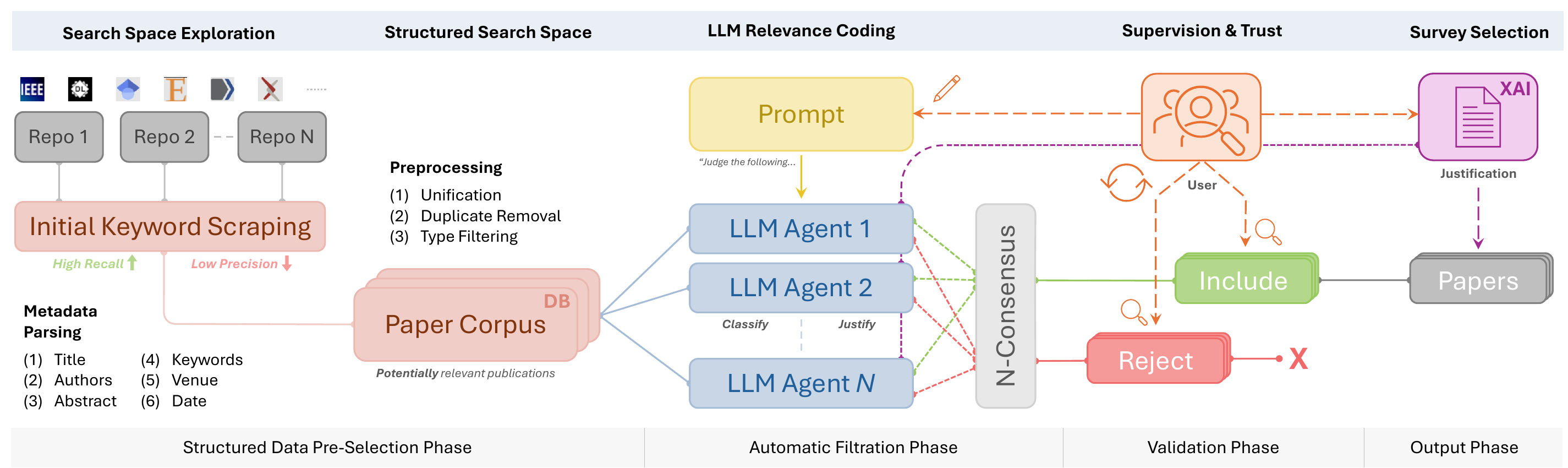}
    \centering
    \caption{Schematic overview of leveraging LLM-based agents for structured literature filtration in a systematic review (SLR). Keyword-based search in online libraries generates a large set of candidate papers that are classified by multiple LLMs based on title and abstract using a customized prompt. A consensus voting scheme determines inclusion or rejection, providing justifications that users can review and refine.
    }
   \label{fig:teaser}
}

\maketitle
\begin{abstract}
Systematic literature reviews (SLRs) are essential but labor-intensive due to high publication volumes and inefficient keyword-based filtering. To streamline this process, we evaluate Large Language Models (LLMs) for enhancing efficiency and accuracy in corpus filtration while minimizing manual effort. Our open-source tool LLMSurver presents a visual interface to utilize LLMs for literature filtration, evaluate the results, and refine queries in an interactive way. We assess the real-world performance of our approach in filtering over 8.3k articles during a recent survey construction, comparing results with human efforts. The findings show that recent LLM models can reduce filtering time from weeks to minutes. A consensus scheme ensures recall rates >98.8\%, surpassing typical human error thresholds and improving selection accuracy. This work advances literature review methodologies and highlights the potential of responsible human-AI collaboration in academic research.


\end{abstract}  

\section{Introduction}
\label{sec:introduction}

Literature reviews, and in particular \emph{systematic literature reviews} (SLRs), have been described as the \textit{gold standard} for conducting literature research in academia~\cite{egger2008systematic, Davis.SystematicReviews.2014}.
They provide a transparent and reproducible approach to systematically synthesize and categorize research findings, providing a comprehensive overview of a research topic~\cite{Nightingale.SLRGuide.2009}.
Such reviews date back to the 18th century~\cite{VanDinter.SLRAutomation.2021} and help identify research gaps, future directions, and ensuring consistency and reliability in academic research~\cite{Lame.SLRIntroduction.2019}.
The creation of SLRs, however, is typically a highly manual and labor-intensive process~\cite{SilvaJunior.RoadmapSLR.2021}.
Egger et al. describe the standard process through a set of eight stages (1. research question, 2. criteria definition, 3. locating, 4. selection, 5. assessment, 6. data extraction, 7. presentation, 8. interpretation)~\cite{egger2008systematic}.
With \textit{PRISMA}~\cite{prisma2009}, a well-established, standardized method exists, describing the process of retrieving a paper corpus (steps 3--6) by keyword-based search, duplicate removal, manual screening based on title and abstract, and the full-text manuscript review.
One of the most time-consuming tasks in this pipeline is the manual title and abstract screening, particularly in domains or research fields where classical keyword-based filtering may lead to ambiguous results.
According to Wallace et al.~\cite{Wallace.SemiAutomatedScreeningSystematicReview.2010}, an experienced peer-reviewer can manually screen about two papers per minute based on title and abstract.
At this rate, a corpus of about 8,000 potentially relevant publications for a large SLR requires approximately 66 person-hours (about one and a half full work weeks) of \emph{uninterrupted} work time.
Effects like fatigue, loss of accuracy, inefficiencies, dual verification, and other work commitments typically increase the required time frame significantly, resulting in survey latency times closer to a few months for the initial screening alone.
Given the repetitive but still demanding nature of the tasks and the ever-faster progress in academia, it stands to reason if--and how--this process can be improved upon.
\indent
Using automation for such a (relatively) well-defined classification task is not a new idea, with the first use of automation being reported in the mid-2000s~\cite{VanDinter.SLRAutomation.2021}.
However, the recent advancements of Large Language Models (LLMs) prove promising for the tasks of initial literature filtration during the creation of SLRs primarily due to two reasons: (1) their capability to understand nuanced semantic ambiguities, potentially reaching a feasible accuracy (recall, precision) threshold, and (2) their unparalleled speed and cost-efficiency w.r.t. to human labor.
While existing LLM chatbots have the ability to search external databases through function calls, limited research has been conducted on a schematic pipeline of the whole \textit{process} from repository acquisition to final paper selection and its \textit{evaluation}, ensuring \textit{completeness}, \textit{reliability}, and \textit{accountability}, which is the focus of this research.
In this work, we present a visual-interactive approach leveraging LLMs for literature filtration during SLR creation, allowing users to iteratively refine prompts, evaluate the results, and interactively create a consensus scheme leading to the desired classification result.
Thereby, we make the following contributions:
\begin{itemize}
\setlength\itemsep{0.2em}
   \item A \textbf{conceptual schema} for the structured literature filtration process leveraging LLM-based agents with consensus voting
	
  \item A visual-interactive open-source \textbf{application}, \href{https://github.com/dbvis-ukon/LLMSurver}{LLMSurver}, implementing our framework and making it accessible to others
	
  \item A comprehensive \textbf{evaluation} for a large SLR (8.3k papers)
  with an extensive \textbf{discussion} on the pitfalls, potentials, and future prospects of leveraging AI agents for literature filtration
\end{itemize}

\section{Related Work}
\label{sec:related_work}

With the recent successes of machine learning and in particular LLMs, an increasing number of publications show how language models can leverage some parts of the tasks~\cite{Wallace.SemiAutomatedScreeningSystematicReview.2010, SilvaJunior.RoadmapSLR.2021} involved in the scientific publishing process~\cite{lund2023chatgpt}.
These tasks include, for instance, generating paper reviews (e.g., to improve the own work)~\cite{tyser2024aidriven}, reformulating paragraphs for clarity~\cite{gilat2023how}, identifying and avoiding biased arguments~\cite{huang2023role}, or finding gaps in previous research for a given domain~\cite{lund2023chatgpt}.
\indent
Besides these general publishing tasks, LLMs have recently been shown to also support various aspects of conducting literature reviews~\cite{Sami.SystematicLiteratureReviewAIAgents.2024}.
While a large body of research on how to conduct literature reviews exists~\cite{Nightingale.SLRGuide.2009, Davis.SystematicReviews.2014, egger2008systematic, Lame.SLRIntroduction.2019}, many of the developed methods are labor-intensive and repetitive.
Therefore, it has been investigated how agent-based systems can help with the formulation, filtering, and search of a research domain~\cite{Whitfield.ElicitAILiteratureReview.2023, Sami.SystematicLiteratureReviewAIAgents.2024, huang2023role}, using LLMs for specific keyword generation and retrieval through RAG~\cite{Agarwal.LitLLM.2024}, or more generally, how machine learning~\cite{Wallace.SemiAutomatedScreeningSystematicReview.2010, VanDinter.SLRAutomation.2021, SilvaJunior.RoadmapSLR.2021}, but also LLMs~\cite{Antu.LLMLiteratureReview.2023, susnjak2023prisma, rathi2023p21, bolanos2024artificial, hawkins2024literature, peinl2024usingLLM} can support the overall process.
Further, the summarization step may be supported using LLMs~\cite{li2024chatcite}.
One particular aspect that has received less attention is the accurate filtering and classification of a (relatively) large body of potentially relevant research concerning a particular research question to speed up the \textit{paper pre-selection process}.
This is particularly relevant for topics or domains where keyword-based filtering is difficult to use, for example, due to semantic ambiguities or duplicated word use.
Haryanto~\cite{haryanto2024llassist} explores the usability of LLMs for performing this specific task, focusing on the vote of individual LLMs.
Also, fairly recently, automatic tooling approaches for SLR-generation using LLMs have been proposed~\cite{scherbakov2024emergence, gana2024leveraging, jafari2024streamlining, susnjak2024automating}.
Gehrmann et al.~\cite{gehrmann2024large} introduced the only LLM-based automated pre-selection approach, showing that negative prompting can boost accuracy.
Building on this, we propose a similar pipeline for classifying large paper corpora, designed as a visual-interactive process that incorporates and evaluates voting schemes from multiple LLM agents and lets users iteratively refine prompts and LLMs.
We also compare results with a manual SLR selection on the same dataset, offering insights into reliability and accuracy.

\section{Methodology}
\label{sec:methodology}

\begin{figure}[b]
\vspace*{-3mm}
    \fbox{
   \begin{minipage}[b]{0.965\linewidth}
   \setlength{\parskip}{2pt}
   \scriptsize \fontfamily{qag} \fontsize{6}{8}\selectfont
    You are a professor in computer science conducting a literature review.
    Please decide and classify if the following paper belongs to a specific research direction or not.
    For this, you are provided with the title and the abstract, which should give you sufficient information for an informed and accurate decision.
    
    The research direction is the topic of "TITLE".
    
    Therefore include papers that deal with ASPECT\_1, ASPECT\_2, ... Examples of ASPECT\_1 are: term 1, term 2, \ldots
    
    You MUST discard papers that EXCLUSION\_EXCEPTION\_1, \ldots
    
    You MUST include papers that INCLUSION\_EXCEPTION\_1, \ldots
    
    Below is the title and abstract. You must only answer with INCLUDE or DISCARD and a 2-sentence reason of why.
    \end{minipage}
    }
    \caption{Prompt template for the individual agents.}
    \label{fig:prompt_format}
    \vspace*{-2mm}
\end{figure}

\begin{figure*}
     \centering
     \begin{subfigure}[b]{0.45\textwidth}
         \centering
         \includegraphics[width=\textwidth]{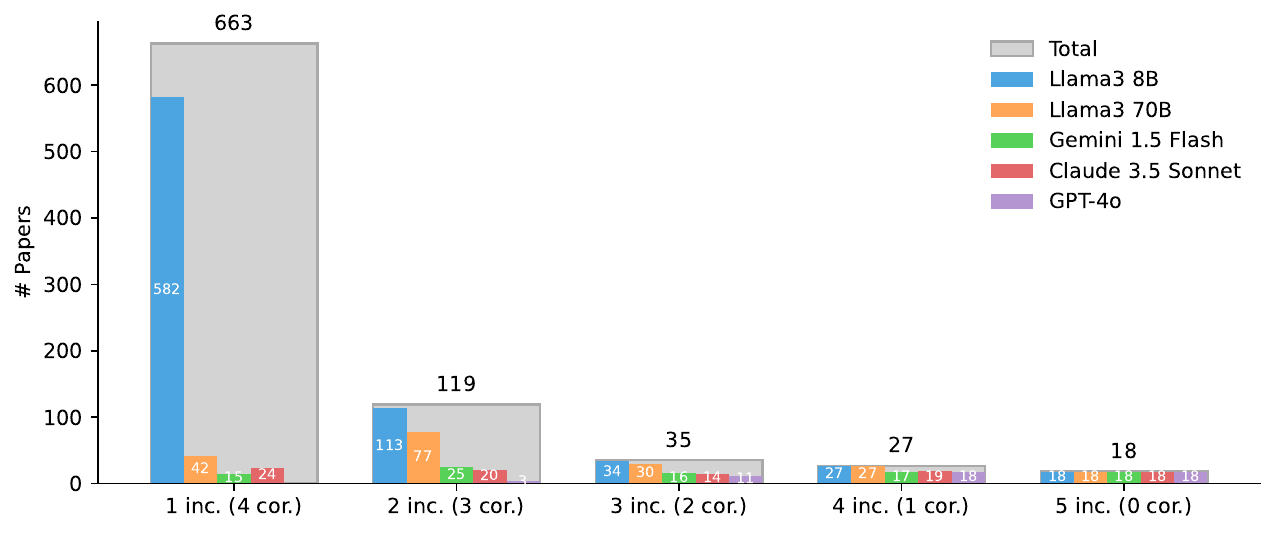}
     \end{subfigure}
     \hfill
     \begin{subfigure}[b]{0.45\textwidth}
         \centering
         \includegraphics[width=\textwidth]{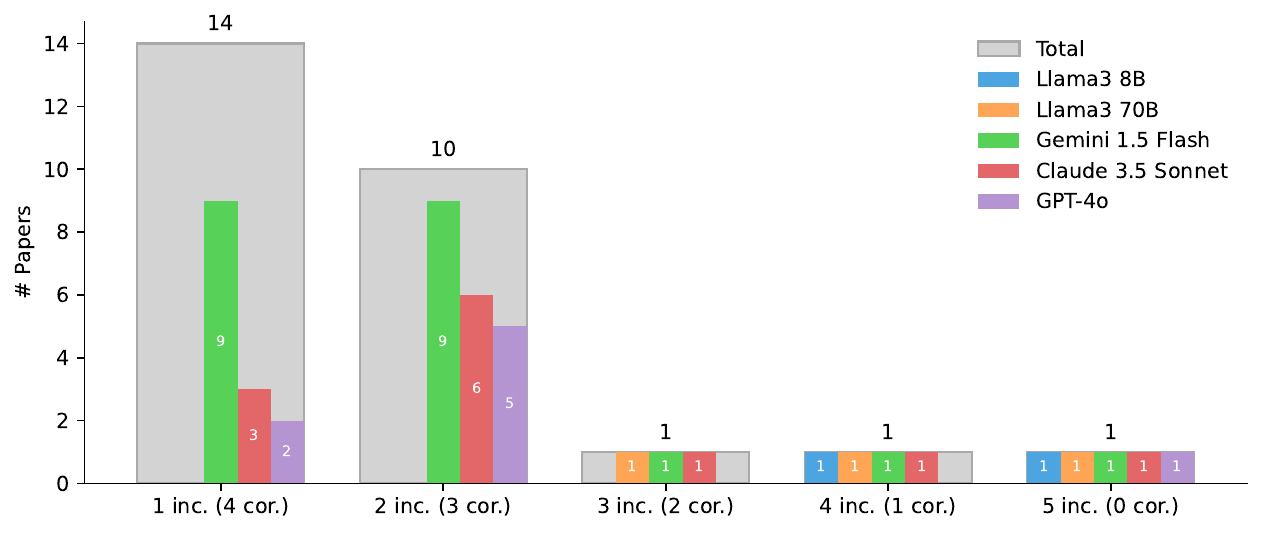}
     \end{subfigure}
        \caption{Number of papers (gray background) that were \emph{incorrectly (inc.)} voted to be \textbf{included} (left) or \textbf{excluded} (right) by the agents, grouped by the number of incorrect agents involved in a decision. The individual bars show how many times a particular agent was involved in the wrong decision. It can be seen on the far right that only a single paper is unanimously misjudged by all agents (and therefore lost forever), demonstrating that N-Consensus voting is beneficial when prioritizing recall.}
        \label{fig:llm-graphs}
        \vspace*{-2mm}
\end{figure*}

To evaluate the applicability of LLMs for pre-filtering the paper corpus for an SLR, we followed a structured methodology, starting with a topic definition, using an early, preliminary version of our recent literature survey
``Visual Network Analysis in Immersive Environments: A Survey''~\cite{joos2025visualnetworkanalysisimmersive}.
This topic leads to a sufficiently large corpus of potential papers since all papers dealing with some immersive technology, such as Virtual Reality, Augmented Reality, and others, focusing on the widespread data type of graphs are of relevance.
For the initial paper selection, we followed the \textit{PRISMA} pipeline~\cite{prisma2009}, starting with a structural keyword-based search in paper titles and abstracts of potential paper candidates in major computer science repositories.
In our case, we included papers from the \textit{ACM Digital Library}, \textit{IEEE Xplore}, and \textit{Eurographics}.
After unifying the format of the paper metadata, removing duplicates, and excluding all non-paper publications, we retrieved an initial corpus of \textbf{8,323} papers in the preliminary version (the published version used a later iteration with more results).
Papers were manually screened in multiple iterations.
This process was highly time-consuming, involving multiple researchers for multiple weeks, but led to the ground truth categorization: for the initial corpus, we identified \textbf{88} papers that needed to be included and \textbf{8235} to discard.
Based on the ground-truth categorization, we investigate the potential for LLMs to facilitate the laborious process of filtering a paper corpus, considering five open and commercial state-of-the-art foundation models (see \autoref{tab:OverallResults} for details).\\
\indent
We initially tested the LLMs with different prompt styles, asking the models to classify each paper individually, and quickly found a basic prompt schema that works well.
In this schema, we tell the LLM its \textbf{context and role}, the \textbf{overall task}, before concluding with an output format and the paper \textbf{title and abstract}.
For the final prompt (see \autoref{fig:prompt_format}), we added further \textbf{exclusion and inclusion criteria}, leading to the following results.

\section{Evaluation}
\label{sec:evaluation}

\begin{table}
\vspace*{-2mm}
\setlength{\tabcolsep}{2.5pt}
\tiny
\centering
\caption{\textbf{Evaluation results} of the LLM agents and two consensus schemes (all models and best models only) for our reference survey, with the validated human classification as ground truth.}
\label{tab:OverallResults}
\begin{tabularx}{\linewidth}{Xr|rrrrr|rr}
\toprule
& \rotatebox{0}{Metric}
& \rotatebox{0}{\makecell{Llama-3\\(8B)$\,^1$}} %
& \rotatebox{0}{\makecell{Llama-3\\(70B)$\,^2$}}
& \rotatebox{0}{\makecell{Gemini 1.5\\Flash$\,^3$}}
& \rotatebox{0}{\makecell{Claude 3.5\\Sonnet$\,^4$}}
& \rotatebox{0}{GPT-4o$\,^5$} 
& \rotatebox{0}{\makecell{Consensus\\(All)$\,^6$}}
& \rotatebox{0}{\makecell{Consensus\\(Best)$\,^7$}} \\
\midrule
\multirow{4}{*}{\rotatebox{90}{Counts}} 
& \textbf{TP}$\;(\uparrow)$ & 86 & 85 & 67 & 76 & 80 & \textbf{87} & \textbf{87}\\
& \textbf{FP}$\;(\downarrow)$ & 774 & 194 & 91 & 95 & \textbf{50} & 862 & 167 \\
& \textbf{TN}$\;(\uparrow)$ & 7461 & 8041 & 8144 & 8140 & \textbf{8185} & 7373 & 8068 \\
& \textbf{FN}$\;(\downarrow)$ & 2 & 3 & 21 & 12 & 8 & \textbf{1} & \textbf{1} \\
\midrule
\multirow{4}{*}{\rotatebox{90}{Evaluation}}
& \textbf{Acc.}$\;(\uparrow)$  & 90.68  & 97.63 & 98.65  & 98.71  & \textbf{99.30}  & 89.63 & 97.98 \\
& \textbf{Prec.}$\;(\uparrow)$  & 10.00  & 30.47 & 42.41  & 44.44  & \textbf{61.54}  & 9.17 & 34.25 \\
& \textbf{Rec.}$\;(\uparrow)$  & 97.73  & 96.59 & 76.14  & 86.36  & 90.91  & \textbf{98.86} & \textbf{98.86} \\
& \textbf{F\textsubscript{1}}$\;(\uparrow)$ & 18.14  & 46.32 & 54.47  & 58.69  & \textbf{73.39}  & 16.78 & 50.88 \\
\bottomrule
\end{tabularx}
\tiny
\raggedright
\vspace*{0.25em} 

$^1$~\texttt{meta-llama-3-8b-instruct.Q8\_0} $\quad$
$^2$~\texttt{meta-llama-3-70b-instruct.Q4\_K\_M} $\quad$
$^3$~\texttt{gemini-1.5-flash-001} $\quad$
$^4$~\texttt{claude-3-5-sonnet@20240620} $\quad$
$^5$~\texttt{gpt-4o-2024-05-13} $\quad$
$^6$~Consensus between all (five) models. $\quad$
$^7$~Consensus between models with 
$F_1 > 50 \%$ (i.e. without Llama3 variants).%
\vspace*{-4mm}
\end{table}

\begin{figure*}
    \vspace*{-2mm}
    \centering
    \includegraphics[width=1\linewidth]{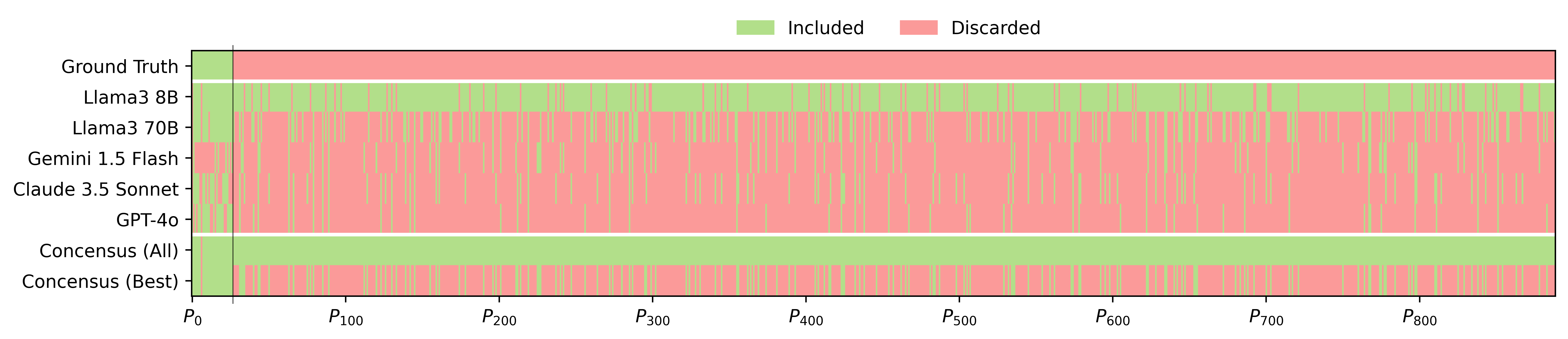}
    \vspace*{-5mm}
    \caption{Pairwise comparison of the \emph{incorrect} decisions by the agents.
    A paper's validated ground truth classification is shown in the top row (left part: included, right part: discarded), followed by the individual agents and then the two consensus methods.
    Incorrect exclusions (FN) can be seen as \textcolor{colorDiscarded}{red discarded} lines (left part), while incorrect inclusion (FP) can be seen as \textcolor{colorIncluded}{green included} lines (larger right part). 
    }
    \label{fig:matrix}
    \vspace*{-4mm}
\end{figure*}

The results of the individual LLM classifications are summarized in \autoref{tab:OverallResults}.
In general, the LLMs performed well with an accuracy above 90 \% across all models.
However, there are still notable differences: While the open-source models--especially Llama3 8B--were more conservative, including more papers in general (high FP rate), trying not to exclude any relevant papers (low FN rate), the commercial models discarded more papers (higher TN rate), but with the downside of having more papers erroneously excluded (higher FN rate).
Interestingly, the falsely classified papers were mostly different across the LLMs:
Regarding the erroneous \textit{inclusions} (FP), for most papers, only one LLM--often Llama3 8B--was responsible for the wrong classification (see \autoref{fig:llm-graphs} left).
The number of papers to exclude that were falsely included by multiple LLMs is drastically lower.
This is also the case for relevant, incorrectly discarded papers (FN), where mostly individual LLMs (mostly Gemini 1.5 Flash) generated errors, but false exclusions by multiple LLMs were way lower (see \autoref{fig:llm-graphs} right).
Therefore, we also analyzed the performance of a consensus voting of all LLMs--Consensus (All)--and a selection of the best-performing LLMs with an $F_1$ score above 50 \%--Consensus (Best)--consisting of Gemini 1.5 Flash, Claude 3.5 Sonnet, and GPT-4o.
For consensus voting, a paper is only discarded if all of the involved LLMs agree to discard it--and included if at least one LLM includes it.
The results of both consensus approaches (see \autoref{tab:OverallResults}, right column) are highly encouraging, showing great results, especially for the \textbf{TP} and the \textbf{FN} rates.
By consensus voting, only one paper would be discarded that should be part of the survey (based on the human ground-truth data).
A manual inspection revealed that this paper was also an edge case for the involved researchers, who might have excluded the paper based on the abstract and title but ultimately included it after investigating its content.
While both consensus approaches lead to the same \textbf{TP} and \textbf{FN} rates, which are of most relevance for our use case, the Consensus (Best) approach comes with a lower \textbf{FP} rate (see \autoref{fig:matrix}), reducing the manual filtering by 695 papers, and only requires three instead of five LLMs, reducing time and cost.

\section{Interactive Human-AI Collaboration}
\label{sec:human-ai}

The results of our experiment show that LLMs can support the initial filtering process.
However, relying solely on (individual) LLMs without human intervention is of high risk.
Therefore, we propose a new pipeline (see \autoref{fig:teaser}) to form the paper corpus of survey papers, incorporating a tight collaboration between the researcher (human) and LLMs (AI).
The first steps of our suggested pipeline remain the same as for classical paper retrieval (see \autoref{sec:methodology}): Online repositories are searched for papers of relevance based on keywords, leading to a pre-processed initial paper corpus.
Then, multiple LLMs classify each paper independently.
The survey authors are an essential part of the process, as they iteratively create and adapt prompts (as in \autoref{fig:prompt_format}) and investigate sampled LLM output through a visual-interactive interface until the results are of sufficient quality.
Investigating the LLMs' justification of their decisions is often highly useful for evaluating the prompt and for refining it.
When the preliminary results are sufficient, all papers are classified by each LLM (or the most promising ones), and the results are combined through a consensus voting.
Again, the consensus results can be iteratively adapted and evaluated through a visual-interactive process, highlighting similarities and differences across the LLMs (similar to \autoref{fig:llm-graphs}).
As our evaluation demonstrated, consensus voting is highly effective in reducing the number of papers, while the rate of erroneously removed papers remains very low.
Human classification can similarly result in false exclusions, although these papers can typically be recovered in a subsequent snowballing step~\cite{wohlin2014guidelines}.
Therefore, a small number of removals during the LLM step may be considered acceptable.

\section{Application}
\label{sec:application}

\begin{figure}[b]
    \centering
    \vspace*{-2mm}
    \includegraphics[width=\linewidth]{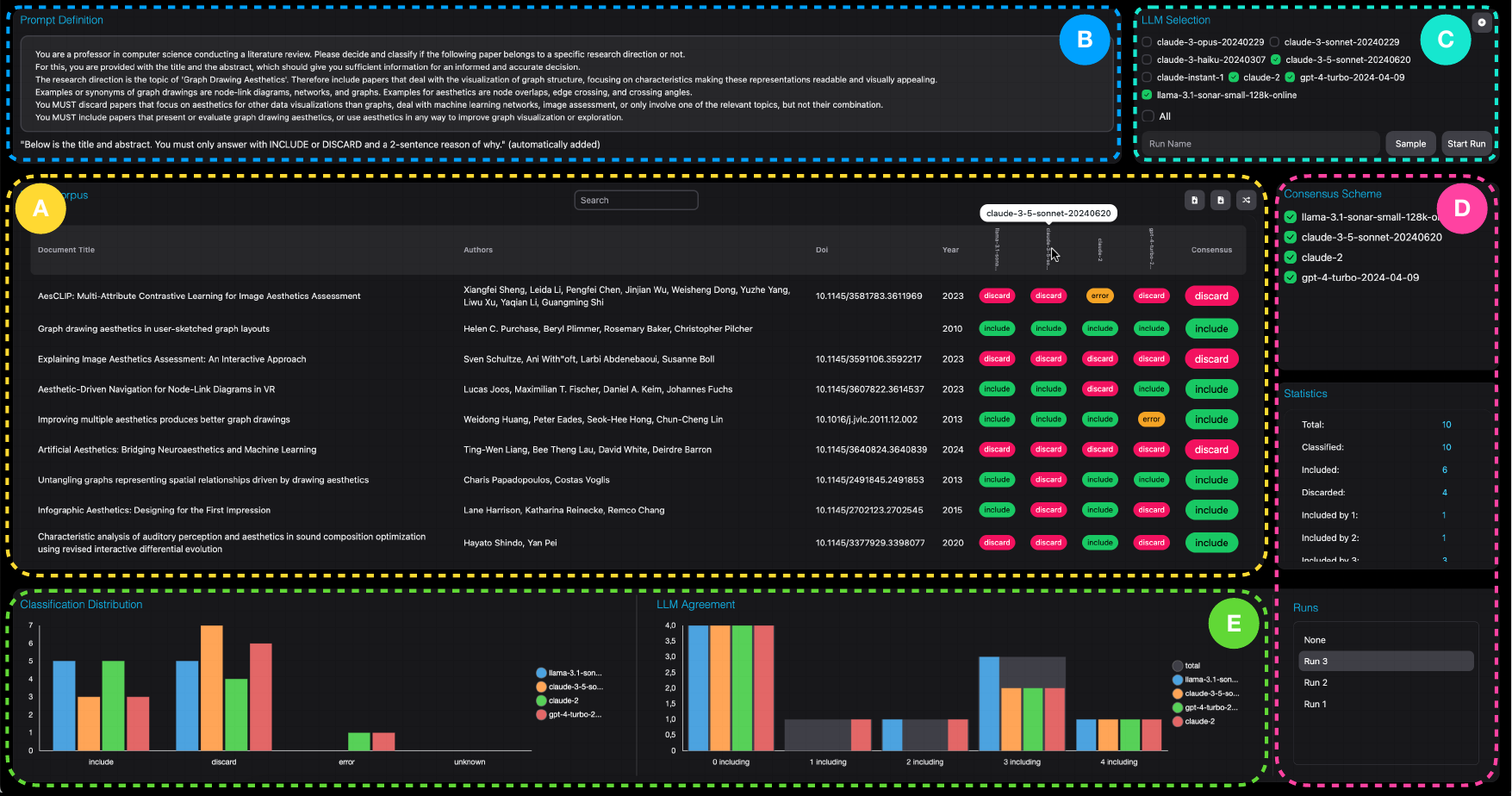}
    \caption{The user interface of our application implementing the proposed pipeline, consisting of a paper table \includegraphics[height=7pt]{img/a.pdf}, a prompt definition area \includegraphics[height=7pt]{img/b.pdf}, a panel for LLM selection classification runs \includegraphics[height=7pt]{img/c.pdf}, the consensus scheme with statistics \includegraphics[height=7pt]{img/d.pdf}, and visual plots \includegraphics[height=7pt]{img/e.pdf}.
    }
    \label{fig:application}
    \vspace*{-2mm}
\end{figure}

We developed the interactive open-source application \textbf{LLMSurver} (\url{https://github.com/dbvis-ukon/LLMSurver}) featuring a user interface (UI) implementing our proposed pipeline. This tool demonstrates the practical application and provides support for researchers conducting their own literature surveys.
The application is fully \texttt{containerized} and follows a frontend (single-page \texttt{React}), backend \texttt{Python}-based \texttt{FastAPI}) database (\texttt{SQLite}) architecture.
The UI is structured as a dashboard, with visually distinct components reflecting the pipeline steps (see~\autoref{fig:application}). The main table \includegraphics[height=7pt]{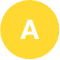} displays paper details from the corpus, populated by uploading \texttt{Bibtex} files or providing \texttt{DOI} numbers.
A prompt editor component \includegraphics[height=7pt]{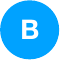} allows users to craft and refine classification prompts for selected LLMs in component \includegraphics[height=7pt]{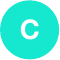}. Users can register new LLMs (local or remote) by entering necessary details such as API keys or hostnames. Classifications can be applied to subsets for testing or the entire corpus, with intermediate results saved.
Classification results are visualized in the main table and can be exported as a \texttt{CSV} file. Users can view individual LLM outputs, particularly useful for ambiguous results (indicated by an orange \texttt{error} icon). 
The consensus component \includegraphics[height=7pt]{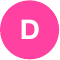} enables run selection, statistics visualization, and LLM selection for consensus-building, with results reflected in the main table. To aid decision-making, component \includegraphics[height=7pt]{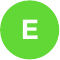} presents two charts: one shows the classification distribution across LLMs, while the other visualizes agreement levels, highlighting outliers (e.g., LLama3 8B in our evaluation) that may reduce consensus quality.
The open-source tool is adaptable for other use cases, supporting custom consensus methods or additional decision-making visualizations.

\section{Discussion}
\label{sec:discussion}

We have demonstrated that incorporating AI techniques, particularly LLM-based agents, into a structured analysis pipeline can effectively support the initial stages of a systematic literature review with surprisingly high quality.
A key advantage is the \textbf{speed and cost-efficiency} of filtration. In our case study with 8,323 papers, GPT-4o processed 4,432,169 input tokens—approximately 532 tokens per paper (including prompts)—and 443,735 output tokens, or around 53 per paper. This entire process was completed in under 10 minutes for just \$28.81 (as of July 2024), demonstrating the \textbf{scalability} of LLMs. Their ability to operate continuously or scale up through additional GPU resources makes them ideal for large-scale literature reviews. Compared to manual filtering, which requires a minimum of 69 hours of concentrated human effort, this represents a significant reduction in both time and cost, making systematic reviews more accessible and efficient.
When \textbf{cost or confidentiality} is a concern, smaller, open-source models--capable of running locally on a standard laptop--still achieve impressive recall rates of 97.73\%, though with lower precision. Even so, these models allow researchers to \textbf{explore research fields} quickly, reducing the manual search space by nearly 90\% (from 8,323 papers to 860) in just a few hours.
Notably, the \textbf{recall difference} between top-performing models and Llama3 8B was minimal, with only one additional paper lost as a false positive. The model's bias towards inclusion requires further investigation. To enhance precision, using a consensus approach reduced false positives from 774 to 167--a 98\% reduction in the validation space, missing only one out of 88.
These results are comparable to \textbf{human error thresholds}, which vary depending on factors like task difficulty, familiarity, stress, and repetition~\cite{Smith.HumanErrorRates.2011}. Established frameworks such as HEART~\cite{humphreys1988human}, TESEO~\cite{bello1980human}, and THERP~\cite{kirwan1988comparative} suggest error rates ranging from 0.5\% to 9\%, placing our filtration results within or even below these ranges. Additionally, during our manual review process, 34 papers were reclassified after initial human filtration, further highlighting the strengths of our LLM-based approach.\\
\indent
A significant advantage of using LLMs is the ability to generate \textbf{consistent and descriptive} classification explanations through prompting, a task that would require considerable additional effort from human reviewers. Automating the initial filtration phase also leads to \textbf{better resource allocation}, allowing researchers to focus on higher-level analysis and interpretation, thereby improving overall productivity while reducing fatigue from repetitive tasks.
This efficiency enables researchers to explore a more \textbf{diverse} range of research fields by lowering the entry cost of initial surveys. Additionally, it can help \textbf{identify gaps} in existing literature by semi-automatically gathering relevant publications for broader overviews of specific topics. The multilingual capabilities of LLMs further enhance the \textbf{accessibility of non-English academic literature}, facilitating the inclusion of relevant publications from specialized venues or fields with older literature not available in English.
Finally, automation inherently improves \textbf{data management}. Using a pipeline architecture helps structure large datasets, making the literature easier to navigate compared to manual processes.

\subsection{Limitations and Future Work}
\label{sec:limitations_future_work}

The use of automation and generative models presents several challenges and limitations.
Our study is based on a \textit{single large corpus and prompt}, which may not generalize to other research areas.
Also, our tool has not yet been evaluated in a \textbf{controlled user study}.
A validity risk, in particular when avoiding snowballing~\cite{wohlin2014guidelines}, is a careful selection of the initial set of bibliographical entries and source databases.
Other factors, such as prompt design, corpus characteristics, or writing style, could also influence performance. Nevertheless, given the strong text comprehension abilities of LLMs, their potential for literature filtration remains promising, warranting further investigation into their capabilities and limitations.
LLMs face well-known challenges, including hallucinations, biases from training data, and accuracy concerns. To ensure \textbf{completeness}, we limited the role of generative AI to classification within a structured schema, avoiding direct involvement in the search process and minimizing the risk of generating false references~\cite{hadi2023survey}.
Despite high accuracy rates, these models can still produce misleading outputs, with performance influenced by model quality, prompt formulation, and contextual understanding. Our study did not focus on \textbf{prompt engineering}~\cite{white2023prompt}, which could potentially improve outcomes. While our approach primarily employed \textbf{zero-shot learning} with contextual examples, exploring few-shot learning could further enhance accuracy, albeit with increased token usage.
Inherent biases, originating from training data or the Reinforcement Learning from Human Feedback (RLHF) process~\cite{bai2022training}, can also lead to skewed or incomplete results. Addressing these biases through interactive feedback loops and visual analytics~\cite{Fischer.EthicalAwareness.2022} is essential for ensuring research accuracy. Although state-of-the-art commercial models demonstrate the highest performance, they present \textbf{access limitations} due to cost, availability, and rate restrictions, potentially disadvantaging smaller research groups or independent researchers~\cite{bommasani2021opportunities}. Developing reliable \textbf{evaluation metrics} for LLM-generated literature surveys is an important area for future research.
A potential risk is the \textbf{over-reliance on automation}, which could undermine researchers' critical thinking and analytical skills~\cite{bommasani2021opportunities}. Balancing automation with human oversight~\cite{Fischer.EthicalAwareness.2022} remains essential.\\
\indent
Future research should explore the development of interactive literature review platforms where LLMs assist researchers in a \textbf{collaborative environment}, integrating user feedback mechanisms into the review process.
While our approach facilitates keyword-based paper search, the potential of new LLMs with access to search engines for \textbf{retrieving the paper corpus} (prompt-based) should be investigated.
Extending the use of LLMs to support \textbf{semi-automatic paper coding}--especially when full-text papers are available--could help evaluate the interpretative capabilities of language models more effectively. Additionally, applying this approach to conduct SLRs across \textbf{multiple disciplines} could enable broader, cross-disciplinary analyses, facilitating research efforts that were previously infeasible due to scale or complexity.

\section{Conclusion}
\label{sec:conclusion}
This work evaluates the potential of LLMs to enhance filtration in academic literature reviews. We propose a semi-automated filtration schema for systematic reviews, leveraging recent foundation models--Llama3 (8B and 70B), Gemini 1.5 Flash, Claude 3.5 Sonnet, and GPT-4o--as classification agents to filter large corpora of publications relevant to specific research questions.
Our method addresses the limitations of traditional keyword-based filtering, which often struggles with semantic ambiguities and inconsistent terminology, requiring time-consuming manual checks. 
With our open-source tool \href{https://github.com/dbvis-ukon/LLMSurver}{LLMSurver}, users can iteratively test different prompts and LLMs while interactively evaluating the results.
We assess LLM performance during the construction of a recent literature survey~\cite{joos2025visualnetworkanalysisimmersive}, comparing results against human filtering on a dataset of 8,323 articles.
The findings show that LLMs can drastically accelerate the review process, shrinking search space by an order of magnitude and reducing weeks of effort to minutes, while maintaining recall ($>98\%$), even below typical human error rates. This efficiency not only enhances SLR but also holds promise for broader academic applications.
Overall, this study highlights the effective use of LLMs to streamline academic research.

\section*{Acknowledgements}

The authors gratefully acknowledge financial support by the Federal
Ministry for Economic Affairs and Climate Action (BMWK, grant
No. 03EI1048D) and the Deutsche Forschungsgemeinschaft (DFG) – Project-ID 251654672 – TRR 161.
\bibliographystyle{eg-alpha-doi} 
\bibliography{egbib}

\newcommand{\etalchar}[1]{$^{#1}$}
\begin{thebibliography}{\uppercase{GLAACG24}}

\bibitem[ACR23]{Antu.LLMLiteratureReview.2023}
\textsc{Antu S.~A., Chen H., Richards C.~K.}:
\newblock Using llm to improve efficiency in literature review for
  undergraduate research.

\bibitem[ALCP24]{Agarwal.LitLLM.2024}
\textsc{Agarwal S., Laradji I.~H., Charlin L., Pal C.}:
\newblock Litllm: A toolkit for scientific literature review, 2024.
\newblock \href {http://dx.doi.org/10.48550/ARXIV.2402.01788}
  {\path{doi:10.48550/ARXIV.2402.01788}}.

\bibitem[BC80]{bello1980human}
\textsc{Bello G., Colombari V.}:
\newblock The human factors in risk analyses of process plants: The control
  room operator model ‘teseo’.
\newblock \emph{Reliability engineering 1}, 1 (1980), 3--14.

\bibitem[BHA{\etalchar{*}}22]{bommasani2021opportunities}
\textsc{Bommasani R., Hudson D.~A., Adeli E., Altman R., Arora S., von Arx S.,
  Bernstein M.~S., Bohg J., Bosselut A., Brunskill E., et~al.}:
\newblock On the opportunities and risks of foundation models, 2022.
\newblock \href {http://dx.doi.org/10.48550/arXiv.2108.07258}
  {\path{doi:10.48550/arXiv.2108.07258}}.

\bibitem[BJN{\etalchar{*}}22]{bai2022training}
\textsc{Bai Y., Jones A., Ndousse K., Askell A., et~al.}:
\newblock Training a helpful and harmless assistant with reinforcement learning
  from human feedback, 2022.
\newblock \href {http://dx.doi.org/10.48550/arXiv.2204.05862}
  {\path{doi:10.48550/arXiv.2204.05862}}.

\bibitem[BSOM24]{bolanos2024artificial}
\textsc{Bolanos F., Salatino A., Osborne F., Motta E.}:
\newblock Artificial intelligence for literature reviews: Opportunities and
  challenges.
\newblock \emph{Artificial Intelligence Review 57}, 10 (2024), 259.
\newblock \href {http://dx.doi.org/10.1007/s10462-024-10902-3}
  {\path{doi:10.1007/s10462-024-10902-3}}.

\bibitem[DMBM14]{Davis.SystematicReviews.2014}
\textsc{Davis J., Mengersen K., Bennett S., Mazerolle L.}:
\newblock Viewing systematic reviews and meta-analysis in social research
  through different lenses.
\newblock \emph{SpringerPlus 3} (2014), 1--9.
\newblock \href {http://dx.doi.org/10.1186/2193-1801-3-511}
  {\path{doi:10.1186/2193-1801-3-511}}.

\bibitem[ESA01]{egger2008systematic}
\textsc{Egger M., Smith G.~D., Altman D.}:
\newblock \emph{Systematic reviews in health care: meta-analysis in context}.
\newblock 2001.
\newblock \href {http://dx.doi.org/10.1002/9780470693926}
  {\path{doi:10.1002/9780470693926}}.

\bibitem[FHJ{\etalchar{*}}22]{Fischer.EthicalAwareness.2022}
\textsc{Fischer M.~T., Hirsbrunner S.~D., Jentner W., Miller M., Keim D.~A.,
  Helm P.}:
\newblock Promoting ethical awareness in communication analysis: Investigating
  potentials and limits of visual analytics for intelligence applications.
\newblock In \emph{Proc. FAcct '22} (2022), ACM, pp.~877--889.
\newblock \href {http://dx.doi.org/10.1145/3531146.3533151}
  {\path{doi:10.1145/3531146.3533151}}.

\bibitem[GC23]{gilat2023how}
\textsc{Gilat R., Cole B.~J.}:
\newblock How will artificial intelligence affect scientific writing, reviewing
  and editing? the future is here...
\newblock \emph{Arthroscopy: The Journal of Arthroscopic \& Related Surgery
  39}, 5 (2023), 1119--1120.
\newblock \href {http://dx.doi.org/10.1016/j.arthro.2023.01.014}
  {\path{doi:10.1016/j.arthro.2023.01.014}}.

\bibitem[GLAACG24]{gana2024leveraging}
\textsc{Gana B., Leiva-Araos A., Allende-Cid H., Garc{\'\i}a J.}:
\newblock Leveraging llms for efficient topic reviews.
\newblock \emph{Applied Sciences 14}, 17 (2024), 7675.
\newblock \href {http://dx.doi.org/10.3390/app14177675}
  {\path{doi:10.3390/app14177675}}.

\bibitem[GQB24]{gehrmann2024large}
\textsc{Gehrmann J., Quakulinski L., Beyan O.}:
\newblock Large language models for literature reviews-an exemplary comparison
  of llm-based approaches with manual methods.
\newblock In \emph{Proc. FLLM} (2024), IEEE, pp.~385--391.
\newblock \href {http://dx.doi.org/10.1109/FLLM63129.2024.10852447}
  {\path{doi:10.1109/FLLM63129.2024.10852447}}.

\bibitem[Har24]{haryanto2024llassist}
\textsc{Haryanto C.~Y.}:
\newblock Llassist: Simple tools for automating literature review using large
  language models, 2024.
\newblock \href {http://dx.doi.org/10.48550/arXiv.2407.13993}
  {\path{doi:10.48550/arXiv.2407.13993}}.

\bibitem[HQS{\etalchar{*}}23]{hadi2023survey}
\textsc{Hadi M.~U., Qureshi R., Shah A., Irfan M., et~al.}:
\newblock A survey on large language models: Applications, challenges,
  limitations, and practical usage.
\newblock \emph{Authorea Preprints} (2023).

\bibitem[HT23]{huang2023role}
\textsc{Huang J., Tan M.}:
\newblock The role of chatgpt in scientific communication: writing better
  scientific review articles.
\newblock \emph{AJCR 13}, 4 (2023).

\bibitem[HT24]{hawkins2024literature}
\textsc{Hawkins J., Tivey D.}:
\newblock Efficient systematic reviews: Literature filtering with transformers
  \& transfer learning, 2024.
\newblock \href {http://dx.doi.org/10.48550/arXiv.2405.20354}
  {\path{doi:10.48550/arXiv.2405.20354}}.

\bibitem[Hum88]{humphreys1988human}
\textsc{Humphreys P.}:
\newblock Human reliability assessors guide: an overview.
\newblock \emph{Human factors and decision making: their influence on safety
  and reliability} (1988).

\bibitem[Jaf24]{jafari2024streamlining}
\textsc{Jafari S. M.~A.}:
\newblock Streamlining the selection phase of systematic literature reviews
  (slrs) using ai-enabled gpt-4 assistant api, 2024.
\newblock \href {http://dx.doi.org/10.48550/arXiv.2402.18582}
  {\path{doi:10.48550/arXiv.2402.18582}}.

\bibitem[JFR{\etalchar{*}}25]{joos2025visualnetworkanalysisimmersive}
\textsc{Joos L., Fischer M.~T., Rauscher J., Keim D.~A., Dwyer T., Schreiber
  F., Klein K.}:
\newblock Visual network analysis in immersive environments: A survey, 2025.
\newblock \href {http://dx.doi.org/10.48550/arXiv.2501.08500}
  {\path{doi:10.48550/arXiv.2501.08500}}.

\bibitem[Kir88]{kirwan1988comparative}
\textsc{Kirwan B.}:
\newblock A comparative evaluation of five human reliability assessment
  techniques.
\newblock In \emph{Human Factors and Decision Making: Their influence on safety
  and reliability}. 1988.

\bibitem[Lam19]{Lame.SLRIntroduction.2019}
\textsc{Lame G.}:
\newblock Systematic literature reviews: An introduction.
\newblock In \emph{Proceedings of the design society: Int. conf. on engineering
  design} (2019), pp.~1633--1642.
\newblock \href {http://dx.doi.org/10.1017/dsi.2019.169}
  {\path{doi:10.1017/dsi.2019.169}}.

\bibitem[LAT{\etalchar{*}}09]{prisma2009}
\textsc{Liberati A., Altman D.~G., Tetzlaff J., Mulrow C., G{\o}tzsche P.~C.,
  et~al.}:
\newblock The prisma statement for reporting systematic reviews and
  meta-analyses of studies that evaluate healthcare interventions.
\newblock \emph{BMJ 339} (2009).
\newblock \href {http://dx.doi.org/10.1136/bmj.b2700}
  {\path{doi:10.1136/bmj.b2700}}.

\bibitem[LCL{\etalchar{*}}24]{li2024chatcite}
\textsc{Li Y., Chen L., Liu A., Yu K., Wen L.}:
\newblock Chatcite: Llm agent with human workflow guidance for comparative
  literature summary, 2024.
\newblock \href {http://dx.doi.org/10.48550/arXiv.2403.02574}
  {\path{doi:10.48550/arXiv.2403.02574}}.

\bibitem[LWM{\etalchar{*}}23]{lund2023chatgpt}
\textsc{Lund B.~D., Wang T., Mannuru N.~R., Nie B., Shimray S., Wang Z.}:
\newblock Chatgpt and a new academic reality: Artificial intelligence-written
  research papers and the ethics of the large language models in scholarly
  publishing.
\newblock \emph{J. of the Ass. for Inf. Science and Tech 74}, 5 (2023),
  570--581.
\newblock \href {http://dx.doi.org/10.1002/asi.24750}
  {\path{doi:10.1002/asi.24750}}.

\bibitem[Nig09]{Nightingale.SLRGuide.2009}
\textsc{Nightingale A.}:
\newblock A guide to systematic literature reviews.
\newblock \emph{Surgery (Oxford) 27}, 9 (2009), 381--384.
\newblock Determining surgical efficacy.
\newblock \href {http://dx.doi.org/10.1016/j.mpsur.2009.07.005}
  {\path{doi:10.1016/j.mpsur.2009.07.005}}.

\bibitem[PBH{\etalchar{*}}24]{peinl2024usingLLM}
\textsc{Peinl R., Baernthaler J., Haberl A., Chouguley S.~R., Thalmann S.}:
\newblock Using llms to improve reproducibility of literature reviews.
\newblock \emph{Proc. of 2024 Pre-ICIS SIGDSA Symposium} (2024).

\bibitem[RMBK23]{rathi2023p21}
\textsc{Rathi H., Malik A., Behera D., Kamboj G.}:
\newblock P21 a comparative analysis of large language models (llm) utilised in
  systematic literature review.
\newblock \emph{Value in Health 26}, 12 (2023), S6.
\newblock \href {http://dx.doi.org/10.1016/j.jval.2023.09.030}
  {\path{doi:10.1016/j.jval.2023.09.030}}.

\bibitem[SHJ{\etalchar{*}}24]{scherbakov2024emergence}
\textsc{Scherbakov D., Hubig N., Jansari V., Bakumenko A., Lenert L.~A.}:
\newblock The emergence of large language models (llm) as a tool in literature
  reviews: an llm automated systematic review, 2024.
\newblock \href {http://dx.doi.org/10.48550/arXiv.2409.04600}
  {\path{doi:10.48550/arXiv.2409.04600}}.

\bibitem[SHR{\etalchar{*}}25]{susnjak2024automating}
\textsc{Susnjak T., Hwang P., Reyes N.~H., Barczak A. L.~C., McIntosh T.~R.,
  Ranathunga S.}:
\newblock Automating research synthesis with domain-specific large language
  model fine-tuning.
\newblock \emph{ACM Trans. Knowl. Discov. Data} (2025).
\newblock \href {http://dx.doi.org/10.1145/3715964}
  {\path{doi:10.1145/3715964}}.

\bibitem[SJD21]{SilvaJunior.RoadmapSLR.2021}
\textsc{Silva~Júnior E. M.~d., Dutra M.~L.}:
\newblock A roadmap toward the automatic composition of systematic literature
  reviews.
\newblock \emph{IJSMC 1}, 2 (2021), 1–22.
\newblock \href {http://dx.doi.org/10.47909/ijsmc.52}
  {\path{doi:10.47909/ijsmc.52}}.

\bibitem[SR{\etalchar{*}}24]{Sami.SystematicLiteratureReviewAIAgents.2024}
\textsc{Sami A.~M., Rasheed Z., et~al.}:
\newblock System for systematic literature review using multiple ai agents:
  Concept and an empirical evaluation, 2024.
\newblock \href {http://dx.doi.org/10.48550/ARXIV.2403.08399}
  {\path{doi:10.48550/ARXIV.2403.08399}}.

\bibitem[SS11]{Smith.HumanErrorRates.2011}
\textsc{Smith D.~J., Simpson K.~G.}:
\newblock Chapter 5 - reliability modeling techniques.
\newblock In \emph{Safety Critical Systems Handbook}. 2011, pp.~89--106.
\newblock \href {http://dx.doi.org/10.1016/B978-0-08-096781-3.10005-7}
  {\path{doi:10.1016/B978-0-08-096781-3.10005-7}}.

\bibitem[Sus23]{susnjak2023prisma}
\textsc{Susnjak T.}:
\newblock Prisma-dfllm: An extension of prisma for systematic literature
  reviews using domain-specific finetuned large language models, 2023.
\newblock \href {http://dx.doi.org/10.48550/arXiv.2306.14905}
  {\path{doi:10.48550/arXiv.2306.14905}}.

\bibitem[TSL{\etalchar{*}}24]{tyser2024aidriven}
\textsc{Tyser K., Segev B., Longhitano G., et~al.}:
\newblock Ai-driven review systems: Evaluating llms in scalable and bias-aware
  academic reviews, 2024.
\newblock \href {http://dx.doi.org/10.48550/arXiv.2408.10365}
  {\path{doi:10.48550/arXiv.2408.10365}}.

\bibitem[vTC21]{VanDinter.SLRAutomation.2021}
\textsc{{van Dinter} R., Tekinerdogan B., Catal C.}:
\newblock Automation of systematic literature reviews: A systematic literature
  review.
\newblock \emph{IST 136} (2021).
\newblock \href {http://dx.doi.org/10.1016/j.infsof.2021.106589}
  {\path{doi:10.1016/j.infsof.2021.106589}}.

\bibitem[WFH{\etalchar{*}}23]{white2023prompt}
\textsc{White J., Fu Q., Hays S., Sandborn M., Olea C., et~al.}:
\newblock A prompt pattern catalog to enhance prompt engineering with chatgpt,
  2023.
\newblock \href {http://dx.doi.org/10.48550/arXiv.2302.11382}
  {\path{doi:10.48550/arXiv.2302.11382}}.

\bibitem[WH23]{Whitfield.ElicitAILiteratureReview.2023}
\textsc{Whitfield S., Hofmann M.~A.}:
\newblock Elicit: Ai literature review research assistant.
\newblock \emph{Public Services Quarterly 19}, 3 (2023), 201--207.
\newblock \href {http://dx.doi.org/10.1080/15228959.2023.2224125}
  {\path{doi:10.1080/15228959.2023.2224125}}.

\bibitem[Woh14]{wohlin2014guidelines}
\textsc{Wohlin C.}:
\newblock Guidelines for snowballing in systematic literature studies and a
  replication in software engineering.
\newblock In \emph{Proc. EASE} (2014), ACM.
\newblock \href {http://dx.doi.org/10.1145/2601248.2601268}
  {\path{doi:10.1145/2601248.2601268}}.

\bibitem[WTL{\etalchar{*}}10]{Wallace.SemiAutomatedScreeningSystematicReview.2010}
\textsc{Wallace B.~C., Trikalinos T.~A., Lau J., Brodley C., Schmid C.~H.}:
\newblock Semi-automated screening of biomedical citations for systematic
  reviews.
\newblock \emph{BMC bioinformatics 11} (2010), 1--11.
\newblock \href {http://dx.doi.org/10.1186/1471-2105-11-55}
  {\path{doi:10.1186/1471-2105-11-55}}.

\end{thebibliography}

\end{document}